\documentclass{article}

\usepackage{adjustbox}

\usepackage{amssymb}
\usepackage{amsmath} 
\usepackage{booktabs}
\usepackage{stfloats}
\usepackage{multirow}
\usepackage{graphicx}
\usepackage{changepage}
\usepackage{bbm}

\usepackage[activate={true,nocompatibility},final,tracking=true,kerning=true,spacing=true,factor=1100,stretch=10,shrink=9]{microtype}
\usepackage[compact]{titlesec}      
\usepackage[preprint, nonatbib]{neurips_2023}

\usepackage[utf8]{inputenc} 
\usepackage[T1]{fontenc}    
\usepackage{hyperref}       
\usepackage{url}            
\usepackage{booktabs}       
\usepackage{amsfonts}       
\usepackage{nicefrac}       
\usepackage{microtype}      
\usepackage{xcolor}         

\title{Multi-Resolution Audio-Visual Feature Fusion for Temporal Action Localization}

\author{%
  Edward Fish\\
  University of Surrey\\
  \texttt{edward.fish@surrey.ac.uk} \\
  \And
  Jon Weinbren\\
  University of Surrey \\
  \texttt{j.weinbren@surrey.ac.uk}\\
  \AND
  Andrew Gilbert\\
  University of Surrey \\
  \texttt{a.gilbert@surrey.ac.uk}\\
}

\begin{document}

\maketitle
\begin{abstract}
Temporal Action Localization (TAL) aims to identify actions' start, end, and class labels in untrimmed videos. While recent advancements using transformer networks and Feature Pyramid Networks (FPN) have enhanced visual feature recognition in TAL tasks, less progress has been made in the integration of audio features into such frameworks. This paper introduces the Multi-Resolution Audio-Visual Feature Fusion (MRAV-FF), an innovative method to merge audio-visual data across different temporal resolutions. Central to our approach is a hierarchical gated cross-attention mechanism, which discerningly weighs the importance of audio information at diverse temporal scales. Such a technique not only refines the precision of regression boundaries but also bolsters classification confidence. Importantly, MRAV-FF is versatile, making it compatible with existing FPN TAL architectures and offering a significant enhancement in performance when audio data is available.

\end{abstract}

\section{Introduction}

\begin{figure}[h!]
  \centering
  \includegraphics[trim=0cm 2cm 0cm 2cm, clip, width=1\linewidth]{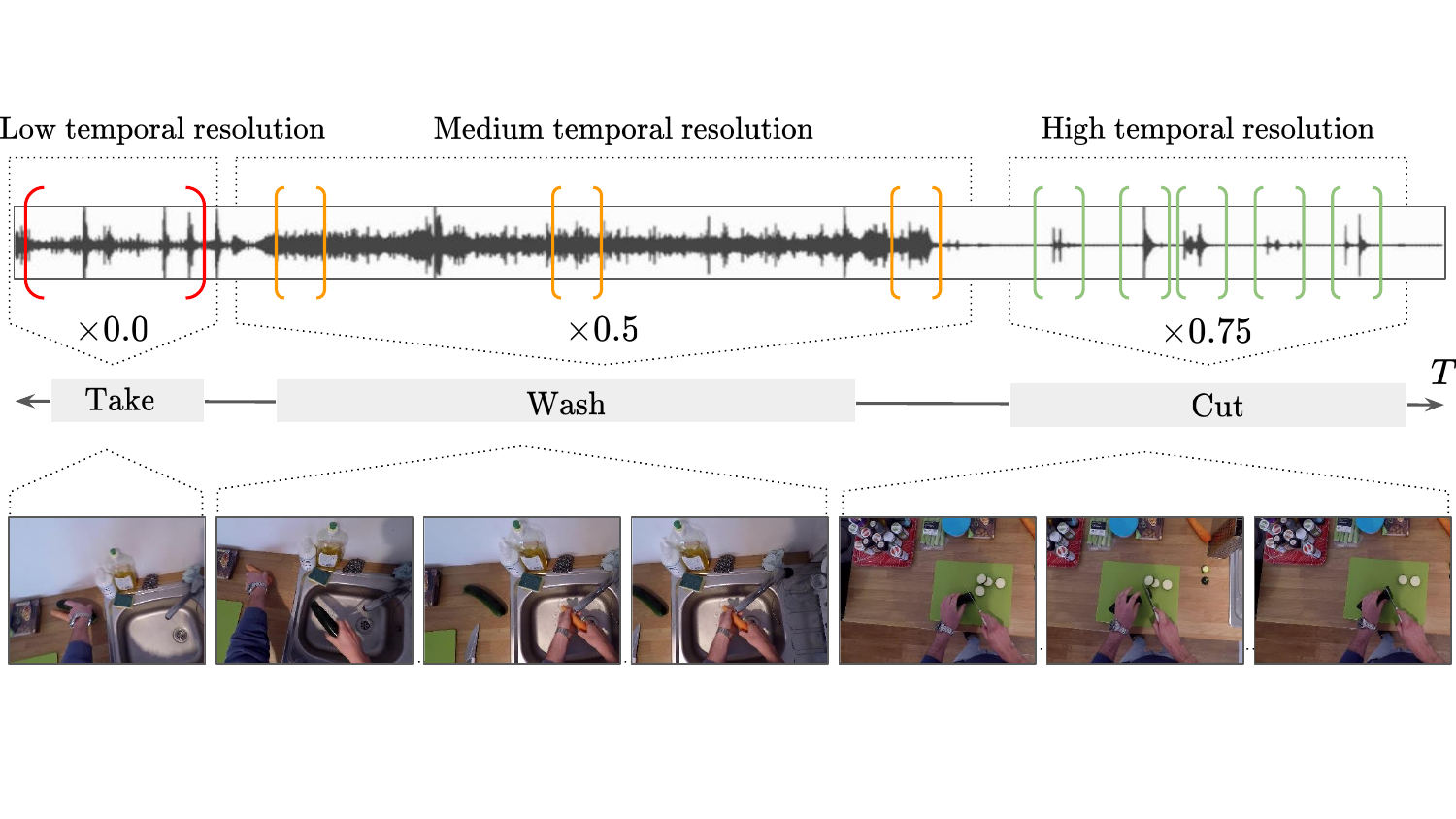}  \caption{We use a Feature Pyramid Network (FPN) to encode audio-visual action features along different temporal resolutions. We then gate the fusion of the audio features depending on their application to the action classification and regression boundaries. For example, the action `take' requires no audio, which is gated out. In contrast, the action `chop' can be better localised by combining high-temporal resolution audio features with visual features. Our method learns both the temporal resolution and the gating values end-to-end.}
  \label{fig:headline}
\end{figure}

Temporal Action Localization (TAL) is concerned with detecting the onset and offset of actions and their class labels in untrimmed and unconstrained videos. Recently, the combined use of transformer networks and Feature Pyramid Networks (FPN)~\cite{yang2020temporal,zhang2022actionformer,cheng2022tallformer,weng2022efficient, shi2023temporal} 
 has led to a significant boost in the performance and efficiency of TAL tasks by leveraging multi-resolution visual features. However, there has not yet been a study on combining audio information in such network architectures for this task, specifically how to fuse audio information over different temporal resolutions. The challenge lies in integrating audio and visual data and determining the density of audio information required across different FPN channels for different actions. While some channels might require richer audio input to accurately identify action segments due to higher visual downsampling, others with more detailed visual cues might need less audio assistance. For instance, as shown in Fig \ref{fig:headline}, an action such as `chopping' can be better located using high-resolution (i.e. less downsampled) audio features. In contrast, an activity such as `washing up' may only require some low-resolution audio information. A final example could be for an action such as `pick-up', which requires no audio input. With this in mind, a fusion method for audio TAL should accommodate multiple temporal audio resolutions while also including a mechanism to gate audio information in specific temporal pathways.

This paper presents a novel framework for Multi-Resolution Audio-Visual Feature Fusion (MRAV-FF) as a first step to solving these issues. Our methodology is rooted in a hierarchical gated cross-attention fusion mechanism that adaptively combines audio and visual features over varying temporal scales. Unlike existing techniques, MRAV-FF weighs the significance of each modality's features at various temporal scales to improve the regression boundaries and classification confidence. Furthermore, our method can be easily plugged into any FPN TAL architecture to boost performance when audio information is available.

\section{Related Work}

\textbf{Temporal Action Localization} (TAL). Methods in TAL can be separated as single and two-stage. Where single stage methods generate a large number of proposal segments which are then passed to a classification head\cite{escorcia2016daps, buch2017sst, heilbron2016fast, lin2018bsn, gong2020scale, zhao2020bottom, lin2018bsn,lin2019bmn,lin2020fast,chen2022dcan,escorcia2016daps,liu2021multi}. Single-stage methods include the use of graph neural networks\cite{bai2020boundary, xu2020g, zeng2019graph,xu2020g} and more recently, transformers \cite{wang2021temporal, chang2021augmented, tan2021relaxed}. Recent progress in single-stage TAL has shown improvements over two-stage methods in accuracy and efficiency, combining both action proposal and classification in a single forward pass. Works inspired by object detection \cite{redmon2016you, liu2016ssd}, saliency detection \cite{lin2021learning}, and hierarchical CNN's ~\cite{yang2020revisiting,lin2021learning,yang2022basictad} all combine proposal and classification. Current SOTA methods in TAL utilise transformer-based \cite{vaswani2017attention} feature pyramid networks (FPN's)~\cite{zhang2022actionformer,cheng2022tallformer,weng2022efficient, shi2023temporal}, which combine multi-resolution transformer features with classification and regression heads.

\textbf{Audio-Visual Fusion}. Audio-visual fusion via learned representations has been explored in several video retrieval and classification tasks~\cite{ephrat2018looking,alcazar2021maas,xiao2020audiovisual,wang2020makes,nagrani2021attention,kazakos2019epic,kazakos2021little,kazakos2019epic}.
Audio-visual TAL has been less explored, with most approaches focused on audio-visual events in which the audio and visual events are closely aligned ~\cite{tian2018audio, bagchi2021hear}. Concurrent works exploring audio-visual fusion in TAL have adopted two-stage late fusion approaches. Recent works have also explored audio-visual cross-attention ~\cite{ramazanova2023owl} but over a single temporal resolution and without any gated fusion control.

\section{Method}

\begin{figure}[h!]
  \centering
   \includegraphics[trim=0cm 0cm 0cm 0.5cm, clip, width=0.9\linewidth]{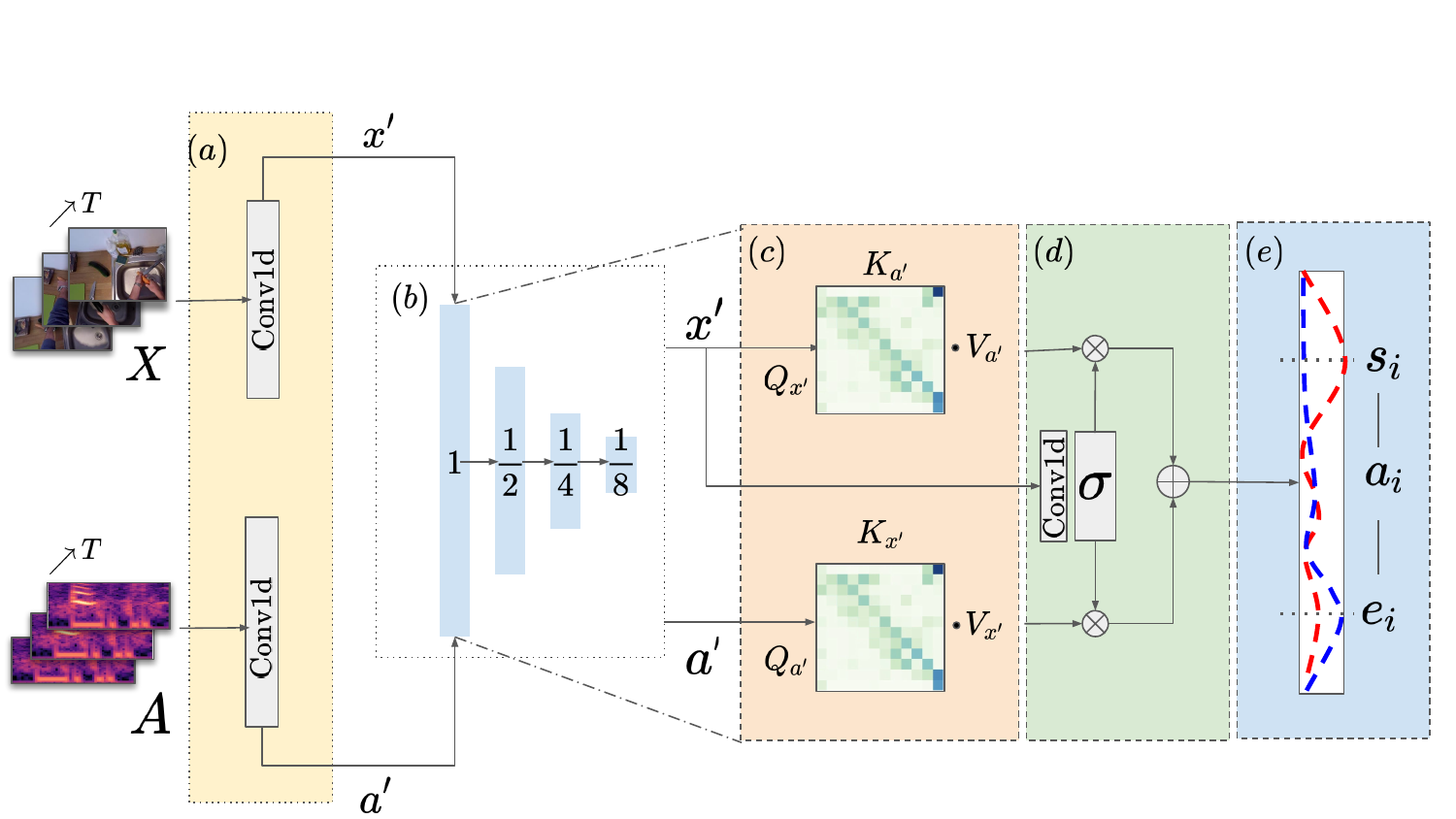}
  \caption{A high-level representation of our multi-resolution audio-fusion method. (a): Audio and visual features are projected to a shared dimension via a 1D convolution. (b) Max-Pooling is applied to downsample features. (c) Following downsampling, we apply multi-headed cross attention in each temporal layer between audio and visual features. (d) The video features are then used as context to scale audio and visual attended embeddings. (e) The concatenated embedding is then used for both regression and classification.}
  \label{fig:cross-att}
\end{figure}

\textbf{Problem Definition} Consider an untrimmed input video denoted as \( \mathcal{X} \). The goal is to represent \( \mathcal{X} \) as a set of feature vectors symbolized as \( \mathcal{X} = \{ x_1, x_2, \dots, x_T \} \). Each \( x_t \) corresponds to discrete time steps, \( t = \{ 1, 2, \dots, T \} \). Notably, the total duration \( T \) is not constant and may differ across videos. For illustrative purposes, \( x_t \) can be envisaged as a feature vector extracted from a 3D convolutional network at a specific time \( t \) within the video. The primary objective of TAL is to identify and label action instances present in the input video sequence \( \mathcal{X} \). These instances are collectively denoted as \( \mathcal{Y} = \{ y_1, y_2, \dots, y_N \} \), where \( N \) signifies the total number of action instances in a given video. This value can be variable across different videos. Each action instance, \( y_i \), is defined by the tuple \( y_i = (s_i, e_i, a_i) \), where \( s_i \) represents the starting time or onset of the action instance, \( e_i \) denotes the ending time or offset of the action instance, and \( a_i \) specifies the action category or label.

The parameters must adhere to the conditions: \( s_i, e_i \in \{ 1, \dots, T \} \), \( a_i \in \{ 1, \dots, C \} \) (with \( C \) indicating the total number of predefined action categories), and \( s_i < e_i \), which ensures the starting time precedes the ending time for every action instance.
Furthermore, alongside the visual feature set \( \mathcal{X} \), we introduce an audio feature set \( \mathcal{A} \). This set can be represented as \( \mathcal{A} = \{ a_1, a_2, \dots, a_{T_{\text{audio}}} \} \), spanning up to \( T_{\text{audio}} \) time steps. Notably, the total duration \( T_{\text{audio}} \) may or may not align with \( T \) from the visual features, depending on the extraction mechanism and granularity of the audio features.

A significant challenge in TAL with multi-modal inputs is to devise an optimum method for fusing visual and audio features. This fusion aims to leverage complementary information from both modalities, enhancing the robustness and accuracy of action localization and classification. 

\textbf{Method Overview} As depicted in Fig \ref{fig:cross-att}, our proposed method is structured around three core components. First, video and audio features are extracted from untrimmed videos using frozen, pre-trained encoders. These encoders provide a robust foundation for capturing the inherent characteristics of the media without additional training overhead. Post-extraction, these features are further refined via a shallow convolution layer. Subsequently, they are channelled into a feature pyramid network. This network's features experience iterative downsampling and are intricately fused through our novel cross attention mechanism. This mechanism ensures effective alignment and integration of features from diverse modalities and resolutions, facilitating the capture of complex temporal relationships. Finally, upon feature fusion, each temporal feature vector is processed by two dedicated decoders: one for regression, predicting action onsets and offsets, and the other for classification, identifying specific action class labels. This dual-decoder approach ensures accurate temporal localization and semantic identification of each detected action.

\textbf{Audio-Visual Temporal Fusion:} Given projected audio embeddings \( \mathcal{A} = \{ a_1, a_2, \dots, a_{T_{\text{audio}}} \} \) and visual embeddings \( \mathcal{X} = \{ x_1, x_2, \dots, x_T \} \) for each timestep, we can break down the process as follows:

\textbf{Downsampling:} For any feature set \( F \), the downsampled feature \( F' \) is computed as: 
\begin{equation}
F' = \text{MaxPool}(F, \text{stride}=2) 
\end{equation}
\textbf{Multi-Headed Cross Attention:} The attention mechanism can be denoted for any feature \( f \) as:
\begin{equation}\text{Attention}(f) = \text{Softmax}(f Q f^T K) V \end{equation}

where \( Q, K, \) and \( V \) are the learned query, key, and value matrices, respectively. Given the downsampled video feature \( x' \) and audio feature \( a' \), the cross-modal projection for the video as query and audio as query is defined as:

\begin{equation} P_{x} = x' Q_{x} (a' K_{a})^T V_{a} \quad \textrm{and} \quad   P_{a} = a' Q_{a} (x' K_{x})^T V_{x} \end{equation}

where \( Q_{x}, K_{x}, V_{x} \) and \( Q_{a}, K_{a}, V_{a} \) are the respective learned matrices for the video and audio modalities. 

\textbf{Gated Audio-Visual Fusion:} To further refine our fusion process, we introduce a gating mechanism which adaptively scales the contribution of audio and visual features based on the context of the visual content. For each downsampled visual feature \( x' \), we compute a gating scalar \( g \) using a sigmoid function:
\begin{equation} g = \sigma(\text{FC}(x')) \end{equation}
where \( \sigma \) denotes the sigmoid activation function, ensuring \( g \) is in the range \([0, 1]\), and \(\text{FC}\) is a fully connected layer. Using the gating scalar, the cross-modal projections are adjusted as follows:
\begin{align}
P_{x, \text{gated}} &= g \cdot P_{x} \quad & P_{a, \text{gated}} &= (1-g) \cdot P_{a}
\end{align}

The combined feature representation after the gated cross-modal projection is then:
\begin{equation} F_{\text{gated\_combined}} = \text{Conv1D}([P_{x, \text{gated}}; P_{a, \text{gated}}]) \end{equation}

\textbf{Regression and Classification:} Each temporal layer outputs gated features to the classification head and the regression head for action instance detection. The output of each instant $t$ in feature pyramid layer $l$ is denoted as $\hat{o}_{t}^l = (\hat{c}_{t}^l, \hat{d}_{st}^l, \hat{d}_{et}^l)$. 
 
We use the same loss as described in ~\cite{tian2019fcos,zhang2020bridging,zhang2022actionformer}:

\begin{equation} \mathcal{L}=\frac{1}{N_{pos}}\sum_{l,t}\mathbbm{1}_{\{c^l_t>0\}}(\sigma_{IoU}\mathcal{L}_{cls} + \mathcal{L}_{reg}) 
+ \frac{1}{N_{neg}}\sum_{l,t}\mathbbm{1}_{\{c^l_t=0\}}\mathcal{L}_{cls} \end{equation}

Where $\sigma_{IoU}$ is the temporal IoU between the predicted segment and the ground truth action instance, and $\mathcal{L}_{cls}$, $\mathcal{L}_{reg}$ is focal loss~\cite{lin2017focal} and IoU loss~\cite{rezatofighi2019generalized}. $N_{pos}$ and $N_{neg}$ denote the number of positive and negative samples.
The term $\sigma_{IoU}$ is used to re-weight the classification loss at each instant, such that instants with better regression (i.e. of higher quality) contribute more to the training.

\section{Evaluation}

\subsection{Dataset}

\textbf{EPIC-Kitchens 100 \cite{damen2018scaling}} is an egocentric dataset containing two tasks: noun localization (e.g. door) and verb localization (e.g. open the door). It has 495 and 138 videos, with 67,217 and 9,668 action instances for training and inference, respectively. The number of action classes for noun and verb are 300 and 97. We follow all other methods ~\cite{lin2019bmn, zhang2022actionformer,cheng2022tallformer,zeng2019graph, tang2023temporalmaxer}, and report the mean average precision (mAP) at different intersection over union (IoU) thresholds with the average mAP computed over [0.1:0.5:0.1] in Table~\ref{table:sota_epic}. 
\begin{table*}[h]
\centering
\begin{tabular}{cccccccc}
\hline
\multirow{2}{*}{Task} & \multirow{2}{*}{Method}                                    & \multicolumn{6}{c}{tIoU}                                                                      \\ \cline{3-8}
                      &                                                            & 0.1           & 0.2           & 0.3           & 0.4           & 0.5           & Avg           \\ \hline
\multirow{4}{*}{Verb} & BMN \cite{lin2019bmn, damen2020rescaling} & 10.8          & 9.8           & 8.4           & 7.1           & 5.6           & 8.4           \\
                      & G-TAD \cite{xu2020g}                      & 12.1          & 11.0          & 9.4           & 8.1           & 6.5           & 9.4           \\
                      & ActionFormer \cite{zhang2022actionformer} & 26.6          & 25.4          & 24.2          & 22.3          & 19.1          & 23.5          \\
                      & TemporalMaxer \cite{tang2023temporalmaxer}                                        & {27.8} & {26.6} & {25.3} & {23.1} & {19.9} & {24.5} \\ \cline{2-8}
                      & ActionFormer + MRAV-FF & 27.6          & 26.8          & 25.3          & 23.4          & 19.8          & 24.6          \\
                      & TemporalMaxer + MRAV-FF                                        & \textbf{28.5} & \textbf{27.4} & \textbf{26.0} & \textbf{23.7} & \textbf{20.12} & \textbf{25.1} \\ \cline{2-8}
\multirow{4}{*}{Noun} & BMN \cite{lin2019bmn, damen2020rescaling} & 10.3          & 8.3           & 6.2           & 4.5           & 3.4           & 6.5           \\
                      & G-TAD \cite{xu2020g}                      & 11.0          & 10.0          & 8.6           & 7.0           & 5.4           & 8.4           \\
                      & ActionFormer \cite{zhang2022actionformer} & 25.2          & 24.1          & 22.7          & 20.5          & 17.0          & 21.9          \\
                      & TemporalMaxer \cite{tang2023temporalmaxer}                                        & {26.3} & {25.2} & {23.5} & {21.3} & {17.6} & {22.8} \\ \cline{2-8}
                      & ActionFormer + MRAV-FF & 26.4          & 25.4          & 23.6          & 21.2          & 17.4          & 22.8   \\      
                      & TemporalMaxer  + MRAV-FF                                        & \textbf{27.4} & \textbf{26.2} & \textbf{24.4} & \textbf{21.8} & \textbf{17.9} & \textbf{23.5} \\ \hline
\end{tabular}

\caption{The performance of our proposed method on the EPIC-Kitchens 100 dataset. \cite{damen2020rescaling}}
\label{table:sota_epic}
\end{table*}

We show the effectiveness of our audio-fusion method in increasing performance of unimodal models by adding our MRAV-FF to the best performing existing FPN networks. We show how our method improves the performance of both ActionFormer and TemporalMaxer by +0.9 mAP and +0.4 mAP for verbs and +0.9 and +0.7 for nouns.

\subsection{Ablation Results}

We perform initial ablation experiments to evaluate the performance of our proposed method and present the results in Tab \ref{table:ablation_2}. Each experiment is conducted on EPIC-Kitchens, where we edit the temporal fusion method in each temporal block. We first exchange our MRAV-FF temporal block for simple feature fusion in which we concatenate and project the audio-visual features at each temporal scale via a 1D-CNN. We notice that this actually harms network performance over unimodal features demonstrating the need for a gated approach to fusion. Similarly we also replace the block with a max-pooling layer inspired by \cite{tang2023temporalmaxer} which pools channel-wise for feature fusion. Again this method has a negative impact on network performance.

\subsection{Further Results}

Furthermore in Tab \ref{table:competitor} we evaluate our method with other approaches to audio-visual fusion for TAL on EPIC-Kitchens. We show a large increase in performance, which can be attributed to both the effectiveness of the FPN structure for audio visual temporal pooling and also our MRAV-FF fusion module. The lack of available comparative methods for audio-visual fusion further illustrates the importance of updated baselines in this field.

Finally, we also evaluate the method on the THUMOS14 dataset which \cite{idrees2017thumos} contains 200 validation videos and 213 testing videos with 20 action classes. THUMOS14 presents a different challenge to ego-centric audio-visual fusion, since the videos are heavily edited and contain many actions that do not have audio-visual alignment. For example, many videos are of sporting events where there is no localized audio information, contain music, narration, or have no audio at all. Due to these challenges there are no existing TAL audio-visual fusion works to our knowledge that test their methods on THUMOS14.

Following previous work \cite{lin2019bmn, lin2018bsn, xu2020g, zhao2020bottom, zhang2022actionformer}, we trained the model on the validation set and evaluate on the test set.
Our results in Tab \ref{table:thumos} demonstrate that our method struggles to handle this audio-visual disparity only improving on the $0.7$ iou threshold.

\begin{table*}[h]
\centering
\begin{tabular}{cccccccc}
\hline
\multirow{2}{*}{Task} & \multirow{2}{*}{Method}                                    & \multicolumn{6}{c}{tIoU}                                                                      \\ \cline{3-8}
                      &                                                            & 0.1           & 0.2           & 0.3           & 0.4           & 0.5           & Avg           \\ \hline
\multirow{4}{*}{Verb} & Concatenation & 28.02 &	26.96 &	25.5	& 23.48 &	19.87  & 23.89\\
                      & Channel Pooling  & 25.63	 & 24.59	& 23.09	 & 21.14	& 17.95	 & 23.06       \\
                                            & MRAV-FF                                        & \textbf{28.5} & \textbf{27.4} & \textbf{26.0} & \textbf{23.7} & \textbf{20.12} & \textbf{25.1} \\ \cline{2-8}
\multirow{4}{*}{Noun} & Concatenation & 26.39	& 25.42	& 23.57 &	21.19 &	17.42 &	22.8        \\
                      & Channel Pooling  & 25.7 &	24.53	& 22.95	 & 20.52	& 17.04	& 22.21         \\
                      & MRAV-FF                                        & \textbf{27.4} & \textbf{26.2} & \textbf{24.4} & \textbf{21.8} & \textbf{17.9} & \textbf{23.5} \\ 
                      \hline
\end{tabular}

\caption{Results for an ablation experiment on EPIC-Kitchens 100 \cite{damen2020rescaling} TAL task, where we replace the MRAV-FF module with existing approaches to feature fusion including concatenated projection and channel pooling. We observe that simple fusion methods hinder performance when compared with uni-modal FPN networks demonstrating the need for a more nuanced fusion strategy.}
\label{table:ablation_2}
\end{table*}

\renewcommand{\arraystretch}{1.2}
\begin{table*}[ht!]
\centering
\begin{tabular}{cccccccc}
\hline
\multirow{2}{*}{Task} & \multirow{2}{*}{Method}                                    & \multicolumn{6}{c}{tIoU}                                                                      \\ \cline{3-8}
                      &                                                            & 0.1           & 0.2           & 0.3           & 0.4           & 0.5           & Avg           \\ \hline
\multirow{4}{*}{Verb} &    Damen ~\cite{damen2021rescaling}  &  10.83 & 9.84 & 8.43 & 7.11 & 5.58 & 8.36  \\
            & AGT~\cite{nawhal2021activity}  &
                      12.01 & 10.25 & 8.15 & 7.12 & 6.14 & 8.73\\
            & OWL~\cite{ramazanova2023owl}  &
                    14.48 & 13.05 & 11.82 & 10.25 & 8.73 & 11.67 \\
                                            & MRAV-FF                                        & \textbf{28.5} & \textbf{27.4} & \textbf{26.0} & \textbf{23.7} & \textbf{20.12} & \textbf{25.1} \\ \cline{2-8}
\multirow{4}{*}{Noun}  &    Damen ~\cite{damen2021rescaling} & 10.31 & 8.33 & 6.17 & 4.47 & 3.35 & 6.53 \\
            & AGT~\cite{nawhal2021activity} & 11.63 & 9.33 & 7.05 & 6.57 & 3.89 & 7.70  \\
            & OWL~\cite{ramazanova2023owl}  & 17.94 & 15.81 & 14.14 & 12.13 & 9.80 & 13.96  \\                                      
                      & MRAV-FF                                        & \textbf{27.4} & \textbf{26.2} & \textbf{24.4} & \textbf{21.8} & \textbf{17.9} & \textbf{23.5} \\ 
                      \hline
\end{tabular}

\caption{The performance of our proposed method on the EPIC-Kitchens 100 dataset \cite{damen2020rescaling} compared to existing approaches for audio-visual feature fusion on TAL. Our method demonstrates a large increase in performance jointly attributed to the addition of feature pyramid architecture and our fusion strategy.}
\label{table:competitor}
\end{table*}

\begin{table*}[htp!]

\small
\begin{adjustwidth}{-1.8cm}{}
\centering
\begin{tabular}{cccccccccc}
\hline
\multirow{2}{*}{Type}         & \multirow{2}{*}{Model}                                                                   & \multirow{2}{*}{Feature}                         & \multicolumn{6}{c}{tIoU$\uparrow$}         & \multirow{2}{*}{time(ms) $\downarrow$} \\ \cline{4-9} 
                              &                                                                                          &                                                  & 0.3           & 0.4           & 0.5           & 0.6           & 0.7           & Avg.                 \\ \hline
\multirow{14}{*}{Two-Stage}   & BMN \cite{lin2019bmn}                                                   & TSN \cite{wang2016temporal}     & 56.0          & 47.4          & 38.8          & 29.7          & 20.5          & 38.5 & 483* \\
                              & DBG \cite{lin2020fast}                                                  & TSN \cite{wang2016temporal}     & 57.8          & 49.4          & 39.8          & 30.2          & 21.7          & 39.8 & --- \\
                              & G-TAD \cite{xu2020g}                                                    & TSN \cite{wang2016temporal}     & 54.5          & 47.6          & 40.3          & 30.8          & 23.4          & 39.3 & 4440* \\
                              & BC-GNN \cite{bai2020boundary}                                           & TSN \cite{wang2016temporal}     & 57.1          & 49.1          & 40.4          & 31.2          & 23.1          & 40.2 & --- \\
                              & TAL-MR \cite{zhao2020bottom}                                            & I3D \cite{carreira2017quo}      & 53.9          & 50.7          & 45.4          & 38.0          & 28.5          & 43.3 & \textgreater644* \\
                              & P-GCN \cite{zeng2019graph}                                              & I3D \cite{carreira2017quo}      & 63.6          & 57.8          & 49.1          & ---           & ---           & ---  & 7298* \\
                              & P-GCN \cite{zeng2019graph} +TSP \cite{alwassel2021tsp} & R(2+1)1 D \cite{tran2018closer}                  & 69.1          & 63.3          & 53.5          & 40.4          & 26.0          & 50.5 & --- \\
                              & TSA-Net \cite{gong2020scale}                                            & P3D \cite{qiu2017learning}      & 61.2          & 55.9          & 46.9          & 36.1          & 25.2          & 45.1 & --- \\
                              & MUSES \cite{liu2021multi}                                               & I3D \cite{carreira2017quo}      & 68.9          & 64.0          & 56.9          & 46.3          & 31.0          & 53.4 & 2101* \\
                              & TCANet \cite{qing2021temporal}                                          & TSN \cite{wang2016temporal}     & 60.6          & 53.2          & 44.6          & 36.8          & 26.7          & 44.3 & --- \\
                              & BMN-CSA \cite{sridhar2021class}                                         & TSN \cite{wang2016temporal}     & 64.4          & 58.0          & 49.2          & 38.2          & 27.8          & 47.7 & --- \\
                              & ContextLoc \cite{zhu2021enriching}                                      & I3D \cite{carreira2017quo}      & 68.3          & 63.8          & 54.3          & 41.8          & 26.2          & 50.9 & --- \\
                              & VSGN \cite{zhao2021video}                                               & TSN \cite{wang2016temporal}     & 66.7          & 60.4          & 52.4          & 41.0          & 30.4          & 50.2 & --- \\
                              & RTD-Net \cite{tan2021relaxed}                                           & I3D \cite{carreira2017quo}      & 68.3          & 62.3          & 51.9          & 38.8          & 23.7          & 49.0 & \textgreater211* \\ 
                              & Disentangle \cite{zhu2022learning}                                           & I3D \cite{carreira2017quo} & 72.1          & 65.9          & 57.0          & 44.2          & 28.5          & 53.5 & --- \\
                              & SAC \cite{yang2022structured}                                           & I3D \cite{carreira2017quo}      & 69.3          & 64.8          & 57.6          & 47.0          & 31.5          & 54.0 & --- \\
                              \hline
\multirow{7}{*}{Single-Stage} & A²Net \cite{yang2020revisiting}                                         & I3D \cite{carreira2017quo}      & 58.6          & 54.1          & 45.5          & 32.5          & 17.2          & 41.6 & 1554* \\
                              & GTAN \cite{long2019gaussian}                                            & P3D \cite{qiu2017learning}      & 57.8          & 47.2          & 38.8          & ---           & ---           & ---  & --- \\
                              & PBRNet \cite{liu2020progressive}                                        & I3D \cite{carreira2017quo}      & 58.5          & 54.6          & 51.3          & 41.8          & 29.5          & ---  & --- \\
                              & AFSD \cite{lin2021learning}                                             & I3D \cite{carreira2017quo}      & 67.3          & 62.4          & 55.5          & 43.7          & 31.1          & 52.0 & 3245* \\
                              & TAGS \cite{nag2022proposal}                                                 & I3D \cite{carreira2017quo}  & 68.6          & 63.8          & 57.0          &  46.3         & 31.8          &  52.8 & ---\\
                              & HTNet \cite{kang2022htnet}                                                 & I3D \cite{carreira2017quo}   & 71.2          & 67.2          & 61.5          &  51.0         & 39.3          &  58.0 & ---\\
                              & TadTR \cite{liu2022end}                                                 & I3D \cite{carreira2017quo}      & 74.8          & 69.1          & 60.1          & 46.6          & 32.8          & 56.7  & 195* \\
                              & GLFormer \cite{he2022glformer}                                                 & I3D \cite{carreira2017quo}      & 75.9 & 72.6 & 67.2 & 57.2 & 41.8 & 62.9 & ---\\
                              & AMNet \cite{liu2022end}                                                 & I3D \cite{carreira2017quo}      & 76.7          & 73.1          & 66.8          & 57.2          & 42.7          &  63.3 & ---\\
                              & ActionFormer \cite{zhang2022actionformer}                               & I3D \cite{carreira2017quo}      & 82.1          & 77.8          & 71.0          & 59.4          & 43.9          & 66.8 & 80 \\
                              & ActionFormer \cite{zhang2022actionformer} + GAP \cite{nag2022post} & I3D \cite{carreira2017quo}      & 82.3          & ---          & 71.4          & ---       &  44.2          & 66.9  & \textgreater 80 \\
                              & TemporalMaxer & I3D \cite{carreira2017quo}      & \textbf{82.8} & \textbf{78.9} & \textbf{71.8} & \textbf{60.5} & 44.7 & \textbf{67.7} &  \textbf{50} \\ \hline   
                               & TemporalMaxer + MRAVFF & I3D \cite{carreira2017quo} + Audio \cite{hershey2017cnn}       & 82.2 & 78.2 & 71.5 & 59.9 & \textbf{45.3} & 67.4 &  60 \\
                               \hline
                    
\end{tabular}
\end{adjustwidth}

\caption{
Performance of our method on the THUMOS dataset for TAL. We observe that audio-visual fusion on edited videos is much more challenging than the raw-video setting due to the addition of background music, narration, and audio-visual misalignment.}
\label{table:thumos}

\end{table*}

\section{Implementation}

\subsection{Feature Extraction}

\textbf{Visual Features:} We use the features provided by existing works in TAL \cite{zhang2022actionformer, lin2019bmn, xu2020g}. For EPIC-Kitchens features are extracted using a SlowFast network \cite{feichtenhofer2019slowfast} pre-trained on EPIC-Kitchens \cite{damen2020rescaling}. During extraction we use a 32-frame input sequence with a stride of 16 to generate a set of 2304-D features.

\textbf{Audio Features:}
For the audio preprocessing and feature extraction, we followed a series of well-established steps to derive meaningful representations:

\begin{enumerate}
    \item \textbf{Resampling:} All audio data was resampled to a uniform rate of 16 kHz in mono.
    
    \item \textbf{Spectrogram Computation:} We computed the spectrogram by extracting magnitudes from the Short-Time Fourier Transform (STFT). This utilized a window size of 25 ms, a hop size of 10 ms, and a periodic Hann window for the analysis.
    
    \item \textbf{Mel Spectrogram Mapping:} The computed spectrogram was then mapped to a mel scale, producing a mel spectrogram with 64 mel bins that cover the frequency range from 125 Hz to 7500 Hz.
    
    \item \textbf{Log Mel Spectrogram Stabilization:} To enhance the stability and avoid issues with the logarithm function, we calculated a stabilized log mel spectrogram as:
    \[ \text{Log-Mel} = \log(\text{Mel-Spectrogram} + 0.01) \]
    Here, the offset of 0.01 prevents the computation of the logarithm of zero.
    
    \item \textbf{Framing:} Finally, the derived features were segmented into non-overlapping examples spanning 0.96 seconds each. Every example encapsulates 64 mel bands and 96 time frames, with each frame lasting 10 ms.
\end{enumerate}

Following extraction the features are projected to 128-D features via a VGG audio encoder network \cite{hershey2017cnn} pretrained on AudioSet \cite{gemmeke2017audio}. The network outputs embeddings of shape $T \times 128$ where $T$ is the temporal input dimension as defined in the paper.

\section{Conclusion}
We demonstrate an effective method for audio-visual fusion with Feature Pyramid Networks. Our drop-in method can be applied to any FPN architecture for temporal action localization and serves as a competitive benchmark for continued research in audio-visual fusion.

\medskip

\small
\bibliography{neurips_2023}

\end{document}